\pdfoutput=1
\documentclass[11pt]{article}
\usepackage[dvipsnames]{xcolor}
\usepackage{emnlp2023}
\usepackage{times}
\usepackage{latexsym}
\usepackage[normalem]{ulem}
\usepackage[most]{tcolorbox}
\usepackage{booktabs}
\usepackage{graphicx}
\usepackage{multirow}
\usepackage{algorithm}
\usepackage{algorithmicx}
\usepackage{algpseudocode}

\newcommand{\delete}[1]{\{\textit{\sout{#1}}\}}
\newcommand{\sdelete}[1]{\textit{\sout{#1}}}
\newcommand{\colorize}[2]{\colorbox{Salmon!#1!white}{\strut #2}}
\newcommand{\bl}[1]{{\color{blue} #1}}

\usepackage[T1]{fontenc}
\usepackage[utf8]{inputenc}
\usepackage{amsmath}

\usepackage{microtype}

\title{Compressing Context to Enhance Inference Efficiency of\\ Large Language Models}

\author{Yucheng Li\textsuperscript{1}\space, Bo Dong\textsuperscript{1}\space, Chenghua Lin\textsuperscript{2}\thanks{\quad Corresponding author}\space, Frank Guerin\textsuperscript{1}\space \\
\textsuperscript{1}~University of Surrey \quad
\textsuperscript{2}~University of Manchester \\
\texttt{\{yucheng.li, bd00531, f.guerin\}@surrey.ac.uk}\\
\texttt{chenghua.lin@manchester.ac.uk}
}

\begin{document}
\maketitle
\begin{abstract}
Large language models (LLMs) achieved remarkable performance across various tasks. However, they face challenges in managing long documents and extended conversations, due to significantly increased computational requirements, both in memory and inference time, and potential context truncation when the input exceeds the LLM's fixed context length.  This paper proposes a method called \textit{Selective Context} that enhances the inference efficiency of LLMs by identifying and pruning redundancy in the input context to make the input more compact. We test our approach using common data sources requiring long context processing: arXiv papers, news articles, and long conversations, on tasks of summarisation, question answering, and response generation. Experimental results show that Selective Context significantly reduces memory cost and decreases generation latency while maintaining comparable performance compared to that achieved when full context is used.  Specifically, we achieve a 50\% reduction in context cost, resulting in a 36\% reduction in inference memory usage and a 32\% reduction in inference time, while observing only a minor drop of .023 in BERTscore and .038 in faithfulness on four downstream applications, indicating that our method strikes a good balance between efficiency and performance. Code and data are available at \url{https://github.com/liyucheng09/Selective_Context}.
\end{abstract}

\section{Introduction}

Large language models (LLMs) have demonstrated remarkable power and impressive generalisation abilities across a wide range of natural language processing tasks, as well as real-life applications \cite{brown2020language,touvron2023llama,bubeck2023sparks}. However, a major challenge for existing LLMs is processing longer context. Dealing with longer context with LLMs is fundamental in scenarios such as having long conversations, document summarisation, and question answering given long documents. However, it is very computationally expensive, particularly with Transformer based LLMs, due to the quadratic growth of memory and computation associated with the 2-D attention matrix \cite{vaswani2017attention}. This makes LLMs less accessible and sometimes leads to context truncation during inference. Moreover, due to the above limitation, existing LLMs were usually pre-trained with fixed-context windows, which further constrains their capability in processing longer context. 

There are active attempts in reducing the computation and memory cost of the Transformer architecture with sparse attention \cite{child2019generating} or local dense attention \cite{beltagy2020longformer}. There are also efforts to learn  soft prompts with further distillation to save context cost during inference \cite{mu2023learning,chevalier2023adapting}. 
In contrast to existing approaches that primarily focus on 
%This paper, instead of focusing on  
architectures or distillations, we introduce a fresh perspective to tackle the redundancy in the input context itself, thus proposing a complementary, model-agnostic approach that can be potentially combined with other architecture optimisation methods to further enhance inference efficiency.

\begin{figure}[t]
    \centering
    \resizebox{\columnwidth}{!}{
    \fbox{\parbox[c]{1.1\columnwidth}{
        \textbf{Context: } Large Languages Models (LLMs) have shown their ability to perform new tasks,  resulting in a line of work that focuses on further scaling these models. These efforts are based on the assumption \delete{that more parameters will lead to better performance.}

        \vskip 0.1in

        \textbf{Query: } What's the assumption behind the efforts to further scale LLMs?

        \vskip 0.05in

        \textbf{LLMs: } Further scaling Large Language Models will lead to better performance on a wide range of tasks.
    }}}
    \caption{Some context is redundant because LLMs have learned that knowledge. LLMs can generate the correct answer even when these redundancies are deleted.}
    \vskip -0.15in
    \label{fig:example}
\end{figure}

The proposed method is motivated by the potential redundancy and repetition in human language, which has two main sources. The first is the inherent redundancy of natural language. For example, in the conversation \textit{"A: Did you get the chance to pick up groceries today?"}, \textit{"B: Yes, \underline{I did get the groceries.}"},  the underlined part can be seen as a common redundancy in communication. Linguistic studies suggest redundancy is ubiquitous in language \cite{wit1999linguistic}. The other type of input redundancy is from the overlap with training material. As the example in Fig.~\ref{fig:example} shows, if some parts of input have already been included in the pre-training stage of LLMs, then it is safe to delete them and the model can still generate the correct answer. In summary, redundancy in the input context, while beneficial for human comprehension, can be extraneous for LLMs and might lead to unnecessary computational expense.

In this paper, we propose \textit{Selective Context}, which prunes redundant content in a given input context, thereby reducing the computational cost and making better use of the fixed context length in LLMs. \textit{Selective Context} evaluates informativeness of lexical units (i.e., tokens, phrases, or sentences) with self-information \cite{shannon1948mathematical} computed by a base causal language model. By selectively retaining content with higher self-information, our method provides a more compact and efficient context representation for LLMs to process without compromising their performance on various applications.

We evaluate the effectiveness and different settings of \textit{Selective Context} on arXiv papers, BBC News, and real conversation on ShareGPT.com with four NLP tasks: summarisation, question answering, original context reconstruction, and conversation. Experimental results demonstrate that our proposed method can significantly enhance context efficiency of LLMs during inference while maintaining comparable performance compared to that achieved when full context is used.

\section{Self-Information}

Self-information, also known as \textit{surprisal} or \textit{information content}, is a fundamental concept in information theory that quantifies the amount of information conveyed by an event given a distribution~\cite{shannon1948mathematical}. In the context of language modelling, the event can be regarded as one step of generation (i.e., a token) and the distribution corresponds to its output distribution. So the self-information of a token can be defined as the negative log likelihood:
\begin{equation}
I(x) = -\log_2{P(x_t | x_{0}, x_{1}, ... ,x_{t-1})}
\end{equation}
where $I(x)$ represents the self-information of token $x$ and $P(x)$ denotes its output probability. 

In  information theory, self-information measures the level of surprise or uncertainty associated with an event; rare events convey more information and thus have higher self-information, while common events convey less information and have lower self-information. In the context of language modelling, self-information can be used to assess the informativeness of lexical units, e.g., words, phrases, or sentences.
% , to see which pieces of information are more likely to be informativeness or important for understanding the context. 
Lexical units with lower self-information are less informative and thus are more likely to be inferred from the context. As a result, we may treat these parts of input as redundant during LLM inference.

In NLP, self-information has been used to measure surprise in creative language artefacts \cite{bunescu2022distribution}. In addition, related concepts of self-information such as entropy and perplexity are widely used in language model optimisation and evaluation.
\begin{align}
& H(S) = \frac{1}{N}\Sigma_t I(x_t) \\
& PP(S) = 2^{H(S)}
\end{align}
where the entropy $H(S)$ of the sentence $S=(x_0, ..., x_n)$ is the average self-information of words in the sentence, and perplexity $PP(S)$ of the sentence can be calculated with entropy. The property of self-information that is especially relevant to our method is the additivity.
\begin{equation}
    \begin{aligned}
        I(x_0, x_1) &= -\log_2{P(x_0, x_1)} \\
                    &= -\log_2{P(x_0)P(x_1|x_0)} \\
                    &= -\log_2{P(x_0)} - \log_2{P(x_1|x_0)} \\
                    &= I(x_0)I(x_1)
    \end{aligned}
\end{equation}
This means we can calculate the self-information of a lexical unit by simply summing the self-information of the tokens in it.

\begin{figure*}[t]
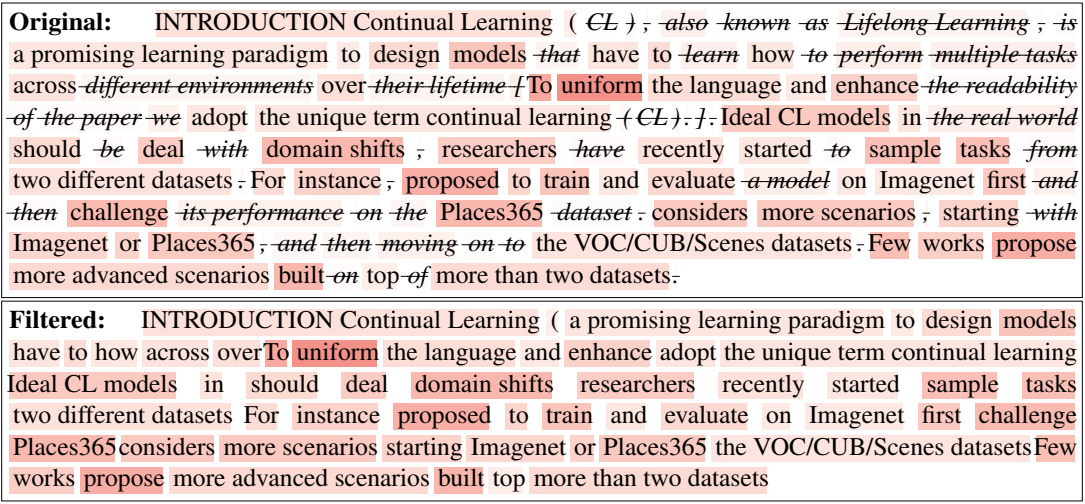

    \centering
\resizebox{0.9\textwidth}{!}{
    \fbox{\parbox[c]{\textwidth}{ 
\textbf{Original: }{\setlength{\fboxsep}{-1pt}
 \input{tex_file/text_0}
}
   \par }}
}
\resizebox{0.9\textwidth}{!}{
\fbox{\parbox[c]{\textwidth}{
\textbf{Filtered: } {\setlength{\fboxsep}{-1pt}
 \input{tex_file/text_1}
}
    }}
}
        \vskip -0.06in
    \caption{A visualisation of selective context. Darker colour indicates larger value of self-information.}
    
    \label{fig:color_text}
        % \vskip -0.06in
    
\end{figure*}

\section{Method}

\textit{Selective Context} optimises the input context by filtering out redundant or non-essential content to reduce computational cost and make better use of the limited context window. 
In implementation, we first 1) employ a causal language model such as GPT \cite{radford2019language,brown2020language}, OPT \cite{zhang2022opt}, or LLaMA \cite{touvron2023llama}, computing self-information  for each token. We then 2) merge tokens, along with their corresponding self-information values, into lexical units, which can be phrases or sentences. This step is optional if tokens are being used as the basic units. Finally, 3) we eliminate content that is deemed least necessary to render the input more compact.

\subsection{Computing Self-Information}

Given a context $C = {x_0, x_1, ..., x_n}$, where $x_i$ denotes a token, we use a base language model $M$ to compute the self-information for each token $x_t$ as follows:
\begin{equation}
I(x_i) = -\log_2{P(x_i | x_0, x_1, ..., x_{i-1})}
\end{equation}
\subsection{Merging into Lexical Units}
\label{merge}

If the content filtering of selective context is directly performed on the token level, it might lead to very disjoint context. Therefore apart from token level filtering, we also conduct the filtering procedure on phrase and sentence level. We call a basic unit in our filtering a \textit{lexical unit}, which could be a token, a phrase or a sentence in our setting.

To enable selective context to work on phrases and sentences, we merge tokens and their self-information into lexical units. Each lexical unit $u$ consists of multiple tokens $(x_t, ..., x_{t+\alpha})$, and we can calculate its self-information by summing the self-information of its individual tokens according to the additivity property of self-information:
\begin{equation}
I(u) = \sum_{i=t}^{\alpha} I(x_i)
\end{equation}
The \texttt{NLTK} sentence tokenizer is employed to obtain sentence level lexical units. And we use \texttt{spacy}\footnote{\url{https://spacy.io/api/pipeline-functions\#merge_noun_chunks}} to merge tokens into noun phrases. We do not merge verb phrases as it might produce very long phrases.

\subsection{Selective Retention of Informative Context}

With the self-information of each lexical unit computed, we can now evaluate their informativeness. Instead of using a fixed threshold or retaining a fixed number of top $k$ lexical units, we design a percentile-based filtering approach to adaptively select the most informative content.

First, we rank the lexical units based on their self-information values in descending order. Then, we compute the $p$-th percentile of self-information values among all lexical units.
\begin{equation}
I_{p} = \texttt{np.percentile}([I(u_0), .., I(u_k)], p)
\end{equation}
Next, we selectively retain lexical units with self-information values greater than or equal to the $p$-th percentile, constructing a filtered context $C'$:
\begin{equation}
C' = {U_i \mid I(U_i) \geq I_{p}, 1 \leq i \leq n}
\end{equation}
The percentile-based filtering is a more flexible approach to retain the most informative content depending on the distribution of self-information values in the given context. In Figure \ref{fig:color_text}, we present an example on phrase level where $p$ is set to 50, which means half of phrases are filtered out. In this case, the context after processing by selective context only retains 57.2\% of tokens, which saves 42.7\% of context length.

\section{Experiments}

The goal of Selective Context is to reduce the redundancy in the input context without compromising the generation quality of LLMs. As a result, we are expecting the answers given both selective context and the original context to be as close as possible. We take the generated answer given full context as the reference answer, and compare to the generated answer given the selective context in our experiments. 

\subsection{Datasets}
Selective Context prunes redundancy in the input context to allow very long context processing for LLMs. However, existing benchmarks for LLMs, such as MMLU \cite{Hendrycks2020MeasuringMM} and ARC \cite{Clark2018ThinkYH}, are mostly single round question answering and are thus not suitable to evaluate our proposed method. Therefore, we collect three test sets consisting of long documents and conversations to evaluate Selective Context. Statistics in detail are presented in Table \ref{tab:dataset_statistics}.

\noindent \textbf{BBC News:} A dataset containing news articles collected from the British Broadcasting Corporation (BBC). This dataset covers a wide range of topics, including politics, business, sports, and technology. We use the full content of each news article in our experiments.

\noindent \textbf{arXiv Articles:} A dataset consisting of latest academic papers, spaning various scientific disciplines, such as computer science, physics, and mathematics. As arXiv articles can be quite long, we only process the first two sections (usually introduction and background) for each paper in our experiments.

\noindent \textbf{ShareGPT.com:} ShareGPT.com is a platform where ChatGPT users share their surprising and interesting conversation with ChatGPT. This datasets consists of conversations in different languages and in various scenarios (e.g., coding, chitchat, writing assistant, etc.). We use the ShareGPT dataset for the conversation task in our experiments.

The three evaluation datasets were created carefully to avoid \textit{data contamination}. Data samples in the BBC News, arXiv, and ShareGPT.com datasets were all created after March 2023, which is after the release of all LLMs in our experiments. Considering some of baseline models are continually being updated, we employ the latest version released before 30 March 2023 to make sure models have never seen our test set in their pre-training and fine-tuning stage. In addition, as some of LLMs in our experiment has a \texttt{max\_length} of 2048 tokens, we do not include articles or conversations exceeding this length.

\subsection{Models}
\label{models}

We test Selective Context on the following models:

\noindent \textbf{GPT-3.5, GPT-4:} GPT-3.5 also known as ChatGPT, which is likely to be further fine-tuned from GPT-3 and InstructGPT. GPT-4 is the latest model from OpenAI, which has demonstrated substantially improved capability on complex reasoning compared to its predecessor. GPT-3.5 and GPT-4 are unfortunately not open-source, we can only access these models via web api\footnote{\url{https://platform.openai.com/docs/api-reference}}.

\noindent \textbf{LLaMA-7, 13, 30B:} LLaMA is a family of open-source language models released by Meta, which is reported to outperform GPT-3 with less parameters. The LLaMA family includes models with size ranging from 7B to 65B. To investigate the effect of scaling law to Selective Context, we experiment with LLaMA with 7B, 13B, and 30B parameters. 

\noindent \textbf{Vicuna-7, 13B:} Vicuna \cite{vicuna2023} is a family of open-source language models instruct-tuned from LLaMA. According to their technical report, Vicuna model perform quite well on a list of multitasking benchmarks.

% There are two main models were used in our experiments: 

% \noindent \textbf{ChatGPT:} We test Selective Context on ChatGPT, which is based on the \texttt{GPT-3.5-turbo} architecture. ChatGPT is a Instruct-tuned language model further improved by RLHF with 175 billion parameters. The base language model of ChatGPT seems to be \texttt{code-davinci-002}\footnote{\url{https://platform.openai.com/docs/model-index-for-researchers}} and more previously \texttt{davinci} which can be found in \cite{brown2020language}. We compare the performance of ChatGPT with and without applying Selective Context to understand its impact on the efficiency and accuracy of the model.

% \noindent \textbf{Curie:} \texttt{Curie} is one of the variant of the GPT-3 family with 6.7B of parameters, a smaller version of causal language model \texttt{davinci}. We employ the Curie as the base model $M$ in Selective Context to calculate self-information. Technically, we shall use the same base model of ChatGPT to do content filtering, but our analysis found that the filtering results of \texttt{curie} and \texttt{davinci} are nearly identical, so for the sake of cost, we choose \texttt{curie} instead.

% We access the two model via web API provided on the OpenAI platform\footnote{\url{https://platform.openai.com/docs/api-reference}}.

\subsection{Tasks and Metrics}

We evaluate Selective Context on four tasks: 

\noindent \textbf{Original Context Reconstruction:} Given a compressed context produced by Selective Context, this task aims to evaluate whether models are able to reconstruct the original context. This task assesses how well the filtered context retains the essential information from the original context. In our experiments, the compressed contexts are used as input, and the original contexts are used as reference answers.

\noindent \textbf{Summarisation:} Given a context, the task is to generate a summary that captures the main points of the document. This task aims to evaluate whether Selective Context affects the overall understanding of models on the input contexts. In our experiments, the input and output are the compressed context and the summaries generated based on the compressed contexts. Summaries based on the \textit{original (full) contexts} are treated as the reference answers.

\noindent \textbf{Question Answering (QA):} Given a document and a set of questions, the task is to generate answers based on the information available in the document. This task aims to evaluate models' understanding towards a specific query. Here we first generate questions and answers based on the original context, where these answers are treated as reference answers, and then ask LLMs to answer these questions with selective context.

\noindent \textbf{Conversation:} This task is only for the ShareGPT dataset. Given a conversation history and a user query, the task is to generate a response to the query based on the previous conversation history. This task aims to evaluate selective context's performance on conversation. Specifically, we ask LLMs to answer users' last query of ShareGPT conversation instances with selective context applied on the previous conversation history. 

We employ four metrics to assess the performance of our models on the tasks: BLEU, METEOR, ROUGE, and BERTScore. BLEU \cite{papineni2002bleu} calculates n-gram precision, which is the proportion of n-grams in the generated text that are also present in the reference text. METEOR \cite{banerjee2005meteor} take additional features such as synonymy, stemming and word order into consideration, which leads to more comprehensive evaluation. ROUGE \cite{lin2004rouge} focuses on how much of the important information in the reference text is present in the generated summary. BERTScore \cite{zhang2019bertscore} leverages contextualised embeddings from pre-trained language models like BERT, computing the cosine similarity between the generated text and reference text embeddings to capture semantic similarity more effectively than traditional n-gram-based metrics.

As mentioned before, we use the generated answers given the full contexts as the reference answers. When testing the deterministic decoding strategy (\texttt{greedy} decoding), we take one single run on full context as the reference answer. When testing non-deterministic decoding strategy (\texttt{temperature = 0.7}), we run multiple times on full context to obtain multiple reference answers to address the randomness in decoding. The metrics are computed based on the set of reference answers. In our experiment, we set the number of reference answers to 4.

\subsection{Experimental Settings}

We use smaller base causal language model for self-information computing in our experiments. For the LLaMA family and vicuna family, we employ LLaMA-7B to compute self-information. For the OpenAI family, we use a smaller GPT-3 variant \texttt{curie} for self-information computing, which is available on OpenAI web API. In self-information computing, we do not process the entire context at once. This is due to our observation on the tendency of LLMs to give later lexical units lower self-information. Instead, we compute self-information sentence by sentence in our experiments. 

In our experiments, we compare the two different dimensions that are adjustable in Selective Context.

\noindent \textbf{Compression Ratios:} We experiment with different content reduction ratios in Selective Context: 0.2, 0.35, 0.5, 0.65, and 0.8. These ratios determine the proportion of content to be filtered out, allowing us to study the trade-off between efficiency and performance as the amount of retained information varies.

\noindent \textbf{Lexical Units: } Lexical units are the basic element of content reduction in Selective Context. It can be sentence, phrases, or tokens. As mentioned in \S\ref{merge}, we remove the redundancy in input context by a specific lexical unit level.
% But due to the usage limitation of OpenAI web API (\$120 per month), we only test the content filtering on phrase level. It doesn't means self-information based content filtering is not feasible on sentence and token level. We will includes experiments on these two level in the next version.

\section{Results}
Except \S\ref{unit}, all results of selective context presented are at the phrase level (the optimal).
\subsection{Overview}

\begin{table*}[t]
\centering
\resizebox{\textwidth}{!}{
\begin{tabular}{llrrrrrrrr}
\toprule
& & & & \multicolumn{3}{c}{ROUGE} & \multicolumn{3}{c}{BERTScore} \\
\cmidrule(lr){5-7}
\cmidrule(lr){8-10}
Method & Ratio & BLEU & METEOR & rouge1 & rouge2 & rougeL & Precision & Recall & F1 \\
\midrule
Original & - & .347 & .496 & .571 & .383 & .471 & .910 & .909 & .909 \\
\cmidrule(lr){2-10}
\multirow[c]{5}{*}{\parbox{2cm}{Selective Context}} & 0.2 & .295 (.05) & .460 (.04) & .540 (\textbf{.03}) & .346 (.04) & .438 (.03) & .905 (.005) & .900 (.009) & .902 (\textbf{.007}) \\
& 0.35 & .243 (.10) & .421 (.08) & .504 (\textbf{.07}) & .294 (.09) & .396 (.07) & .900 (.010) & .894 (.015) & .897 (\textbf{.013}) \\
& 0.5 & .179 (.17) & .362 (.13) & .449 (.12) & .237 (.15) & .344 (.13) & .893 (.018) & .882 (.027) & .887 (.023) \\
& 0.65 & .127 (.22) & .299 (.20) & .391 (.18) & .178 (.21) & .287 (.18) & .885 (.025) & .870 (.039) & .877 (.032) \\
& 0.8 & .070 (.28) & .224 (.27) & .311 (.26) & .122 (.26) & .225 (.25) & .874 (.036) & .852 (.057) & .863 (.047) \\
\bottomrule
\end{tabular}
}
\caption{Comparing Selective Context to the Original Context when \texttt{temperature} set to 0.7.}
\label{tab:to_original}
\end{table*}
\begin{table*}[t]
    \centering
\resizebox{0.9\textwidth}{!}{
\begin{tabular}{lcrrrrrrrr}
\toprule
& & & & \multicolumn{3}{c}{ROUGE} & \multicolumn{3}{c}{BERTScore} \\
\cmidrule(lr){5-7}
\cmidrule(lr){8-10}
& Ratio & BLEU & METEOR & rouge1 & rouge2 & rougeL & Precision & Recall & F1 \\
\midrule
Random & 0.20 & 0.437 &   0.578 &   0.666 &   0.503 &   0.566 &                0.892 &             0.909 &         0.899 \\
                 & 0.35 & 0.360 &   0.514 &   0.629 &   0.423 &   0.502 &                0.879 &             0.895 &         0.886 \\
                 & 0.50 & 0.283 &   0.443 &   0.576 &   0.346 &   0.432 &                0.867 &             0.881 &         0.873 \\
                 & 0.65 & 0.210 &   0.378 &   0.522 &   0.279 &   0.371 &                0.855 &             0.868 &         0.860 \\
                 & 0.80 & 0.156 &   0.314 &   0.450 &   0.219 &   0.310 &                0.843 &             0.853 &         0.847 \\
\midrule
Selective Context & 0.20 & 0.527 &   0.643 &   \textbf{0.714} &   0.585 &   0.631 &                0.930 &             0.932 &         \textbf{0.931} \\
                 & 0.35 & 0.446 &   0.588 &   \textbf{0.679} &   0.508 &   0.570 &                0.915 &             0.916 &         \textbf{0.915} \\
                 & 0.50 & 0.350 &   0.528 &   \textbf{0.642} &   0.425 &   0.501 &                0.900 &             0.902 &         \textbf{0.900} \\
                 & 0.65 & 0.244 &   0.418 &   0.557 &   0.315 &   0.404 &                0.886 &             0.877 &         0.881 \\
                 & 0.80 & 0.160 &   0.328 &   0.464 &   0.223 &   0.319 &                0.875 &             0.858 &         0.866 \\
\bottomrule
\end{tabular}
}
\caption{Comparing Selective Context to the random deletion baseline when using \texttt{greedy} decoding.}
    \label{tab:to_random}
\end{table*}

In Table \ref{tab:to_original}, we first compare the performance of \textit{Selective Context} against the \textit{Original Context} to see how well Selective Context preserves useful information when reducing context cost. The metrics are averaged across all models mentioned in \S\ref{models}. The performance drop is shown in parentheses.

As demonstrated in the table, using Selective Context only leads to a marginal drop when the reduction ratio is set to 0.2 or 0.35, despite it significantly reducing the context cost. The BLEU score drops by only 0.05 when 20\% of the content is reduced. And the number is even smaller when it comes to ROUGE-1, where the drop is just 0.03. This indicate a high level of consistency between answers given selective contexts and original contexts when the reduction ratio is 0.2. Selective Context also yields impressive results when 35\% of the content is reduced, with BERT scores around 0.9 and ROUGE-1 scores over 0.5. The drops become noticeable as the reduction ratio rises to 0.5, where the average BLEU score drops 0.17 and the average ROUGE-1 drops 0.12. A reduction ratio of 0.65 and 0.8 tends to be less valuable, as shown by the 0.18 drop on ROUGE-1 and 0.32 drop on BERTScore-F1.

We then compare \textit{Selective Context} against the \textit{Random} compression baseline as shown in Table \ref{tab:to_random}. We observe that using Selective Context allows LLMs to generate very similar answers to the reference answers (answers given full context) although we significantly reduce the context cost. Selective Context maintains BERTScore-F1 above 0.9 when the compression ratio is 0.5 or lower, which shows a high similarity with the reference answers. ROUGE demonstrates the same trend: ROUGE-1 continue to be above 0.64 and ROUGE-L keeps above 0.5 when the ratio is under 0.5. We also notice that Selective Context is significantly more effective than the random baseline: Selective Context with compression ratio of 0.5 shows a better overlapping with the reference answer than Random baseline with only 20\% content compression.

\subsection{Faithfulness}

\begin{table}[t]
    \centering
    \resizebox{0.8\columnwidth}{!}{
    \begin{tabular}{lrrr}
    \toprule
        Ratio & \#Sorry & Answer len. & Unfaithfulness \\
    \midrule
        Full & 0 & 160.3 & - \\
        \midrule
        0.2 & 0 & 156.5 & .027 \\
        0.35 & 6 & 136.0 & .050 \\
        0.5 & 4 & 140.2 & .038 \\
        0.65 & 19 & 131.2 & .051 \\
        0.8 & 27 & 103.7 & .086 \\
    \bottomrule
    \end{tabular}}
    \caption{Faithfulness test on \texttt{gpt-3.5-turbo} using selective context.}
    \vskip -0.15in
    \label{tab:faithfulness}
\end{table}

\begin{figure*}[t]
    \centering
    \includegraphics[width=0.95\textwidth]{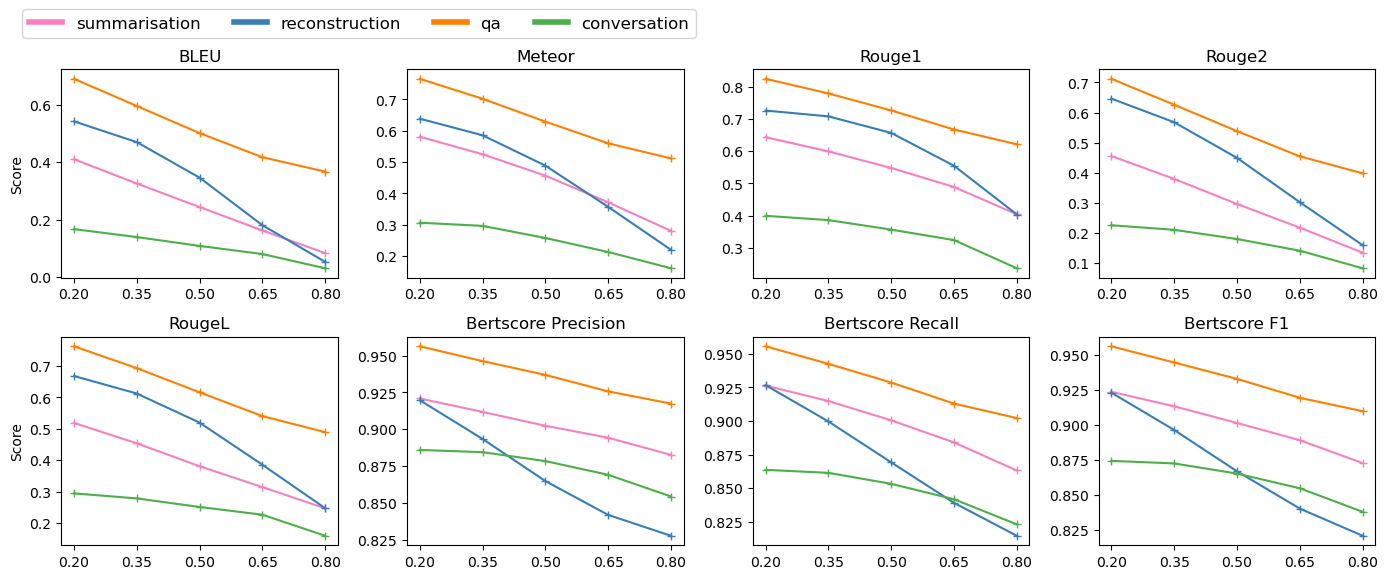}
    \caption{Performance of selective context on different tasks. \texttt{x-axix} represents compression ratios (same below).}
    \vskip -0.15in
    \label{fig:tasks}
\end{figure*}

To evaluate to what extent selective context affects the faithfulness of LLMs generated content, we perform manual tests on our question answering results based on the idea of \cite{Wang2020AskingAA}. We evaluate 1000 question answering pairs (200 for each ratio) with the following procedure: 1) We first extract OpenIE tuples from the answers of selective context, and then 2) manually evaluate whether each tuple is entailed by the reference answer. If the model's answer is "Sorry, I don't know", we treat it as "Sorry" cases and do not consider it as unfaithfulness. 

As shown in the Table \ref{tab:faithfulness}, we find that \texttt{gpt-3.5} tend to generate shorter answers or refuse to answer the questions if it fails to identify necessary evidence in the given selective context. With a compress ratio of 0.65, \texttt{gpt-3.5} refuses to answer 19 questions (9\% of 200), and the answers are 35\% shorter than the reference answer (131 tokens in average). However, selective context doesn't significantly affect the faithfulness across all compression ratios. About 3.8\% of all tuples are not entailed by the reference answer when the compression ratio is 0.5, and this number rises slightly to 5.1\% as the compression ratio increase to 0.65.

\begin{figure}[t]
    \centering
    \includegraphics[width=0.85\columnwidth]{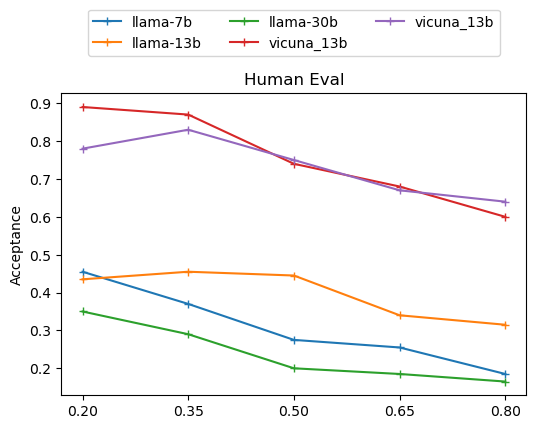}
    \caption{Acceptance rate of generated summaries.}
    \vskip -0.15in
    \label{fig:manual}
\end{figure}

\subsection{Tasks}

In this part, we break down and analyse the performances of Selective Context into the four different NLP tasks: summarisation, question answering, original context reconstruction, and conversation. The results are as shown in Fig.~\ref{fig:tasks}. First, the results on the Original Context Reconstruction task (RC) show the steepest drop with increasing compression ratio, however,  Selective Context  allows LLMs to preserve most of the key points in the original context when the reduction ratio is lower than 0.5, as demonstrated by a rather high ROUGE score. Second, we notice that the curves of question answering and summarisation decrease gradually and are continually higher that the other two tasks evaluated by BERTScore. We could say Selective Context is especially suitable for tasks of summarisation and answer generation.

\begin{figure}[t]
    \centering
\includegraphics[width=0.8\columnwidth]{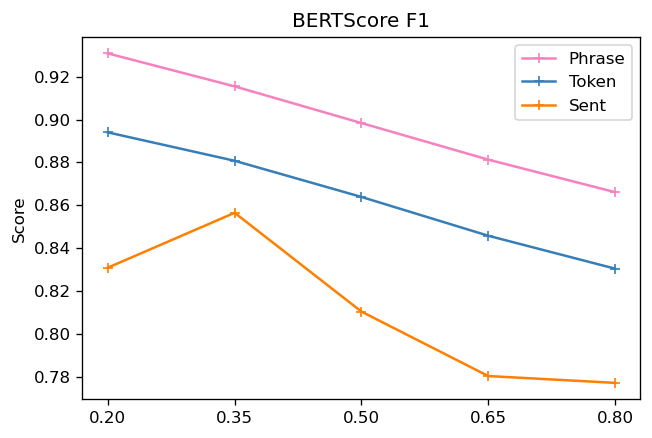}
    \caption{Effects of lexical units.}
    \vskip -0.15in
    \label{fig:units}
\end{figure}

\subsection{Scaling and Instruct-Tuning}

\begin{figure*}[ht]
    \centering
    
    \fbox{\parbox[c]{\textwidth}{
    \textbf{Original Context}, \texttt{CUDA Memory = 77,695 MB; Time = 110.8 ms/token}

    \vspace{0.1in}
        % {\small
        %     \input{tex_file/text_3}
        % }
        Please see the original document and summary given full context in Appendix \ref{sec:appendix}.
    }}
    \fbox{\parbox[c]{\textwidth}{
        \textbf{Selective Context}, Ratio: 0.5, \texttt{CUDA Memory = 61,885 MB, Time = 76.3 ms/token, Time to construct selective context = 46.1 ms}
    \vspace{0.1in}
    
        % {\small
            \input{tex_file/text_4}
        % }
    }}
    \caption{Comparing the summary generated by \texttt{vicuna\_13b} given original context and selective context.}
    % \vskip -0.1in
    \label{fig:case}
\end{figure*}

We perform human evaluation to explore the effect of model scales and supervised instruct-tuning on Selective Context. We ask three college students to evaluate 1150 generated summaries from \texttt{llama} and \texttt{vicuna} (about 55 per model and ratio) and record whether they accept the generation as a reasonable summary.  As shown in Figure \ref{fig:manual}, we find no specific trends between the scales and generation quality given Selective Context. The \texttt{vicuna} family demonstrates similar summarisation capability with \texttt{7b} and \texttt{13b} parameters. And so does the \texttt{llama} family, larger models do not show stronger robustness towards Selective Context. But instruct-tuned model \texttt{vicuna} demonstrates significant superior performance than \texttt{llama} models given selective context indicating instruct-tuning might help the model to be more robustness towards context compression. Given selective context, \texttt{llama} models often fail to follow instructions and go wild very easily.

\subsection{Lexical Units}
\label{unit}

We test the effect of Selective Context based on different lexical units: tokens, phrases, and sentences via BERTScore-F1. As shown in Table \ref{fig:units}, employing phrase as the basic lexical units in Selective Context is the optimal approach, consistently outperforming the other two variants, followed by token-level Selective Context. Removing redundancy at sentence-level is a rather unstable implementation compared to the token and phrase-level. This experiment indicates that a reasonable granularity can be crucial in Selective Context.

\subsection{Case Study}

To have a straightforward impression on how well LLMs generate with selective context, we present two summaries given the full and selective context respectively in Figure \ref{fig:case}. The original document and processing to obtain selective context are presented at Appendix \ref{sec:appendix}.

We first found that preparing selective context is extremely efficient. It takes a one-time cost of 46.1 ms to build selective context for the example paragraph, which includes computing self-information and performing lexical unit tokenisation. This ensures that the initial stage of establishing a selective context incurs very little overhead. Secondly, it shows selective context significantly reduces the memory usage of the GPU and accelerates the generation process. With compression ratio of 0.5, selective context reduces about 36\% of the memory cost in inference and makes  generation 1.32 times faster (per token). By comparing the content of the two summaries, we see that the summary given selective context missed relevant information about the research background (as denoted by the \textsuperscript{\bl{[1]}} marker), such as the use of machine learning in autonomous driving technology and instead starts with the different methods directly. This is due to the background parts not being selected and removed as redundancy before feeding to \texttt{vicuna}. We try to ask \texttt{vicuna} \textit{"what is the background of this study?"} given the selective context, and obtain a decent answer: \textit{"the research background of this paper is likely to be situated in the domain of autonomous driving technology and the application of artificial intelligence (AI) for improving vehicle safety and decision-making capabilities."}. This demonstrates that LLMs are likely to be able to infer the deleted parts of background information in the selective context. Selective context also affects \texttt{vicuna}'s decision on what information should be included in the summary as the second summary includes details about for example FMHSA and UCD block (as denoted by the \textsuperscript{\bl{[2]}} marker) which are not covered in the summary generated with the full context. We find no factual errors in the summary given selective context.

\section{Conclusion}

We introduced Selective Context to improve the context efficiency of LLMs in inference by deleting redundant content measured by self-information. Our extensive experiments on arXiv papers, BBC news articles, and conversation transcripts showed that our proposed method can significantly reduce GPU memory cost, accelerate generation with minor performance decrease, and potentially enable LLMs to handle long documents and extended conversations without the risk of context truncation.

% \clearpage

\section{Limitations}

\textit{Selective Context} demonstrates promising results, but it is still necessary to note a couple of potential limitations. Firstly, our approach is somewhat influenced by the phrase boundary detection procedure. We employ the noun phrase tokenisation algorithm provided by \texttt{spacy} in our experiments. However, we do not consider verb phrases as there is no mature solution for verb phrase tokenisation. We speculate that we can achieve better compression performance with dependency tree-based filtering procedure which might lead to better boundary identification of lexical units. Secondly, in the experiment section, we use percentile to control the pruning process. However, the optimal compression percentile varies based on specific tasks and context. Developing a tool to find the optimal threshold can  further enhance the effectiveness of selective context.
\bibliography{anthology,custom}
\bibliographystyle{acl_natbib}

\appendix

\section{Dataset statistics}

\begin{table}[h]
\centering
\begin{tabular}{lcccc}
\toprule
Dataset & \#Doc & \#Sent & \#Phrase & \#Token \\ \midrule
Arxiv & 408 & 28.20 & 514.55 & 864.85 \\
ShareGPT & 470 & 27.35 & 389.42 & 689.32 \\
BBC & 294 & 25.63 & 523.96 & 732.54 \\ \bottomrule
\end{tabular}
\caption{Statistics of the three datasets. \#Sent, \#Phrase, \#Token are averaged per document.}
\label{tab:dataset_statistics}
\end{table}

\section{Example of selective context on long context}
\label{sec:appendix}

Here we present an example of selective context on a rather long context. The original paragraphs is from \url{https://arxiv.org/abs/2303.07352}. The original paragraphs is shown in Fig. \ref{fig:appendix-case}. The resulting context is shown in Fig. \ref{fig:appendix-resulting}. The reference summary is given in Fig. \ref{appendix-reference-summary}.

\begin{figure*}
    \centering
    \fbox{\parbox[c]{\textwidth}{
    {\setlength{\fboxsep}{-1pt}
    \input{tex_file/text_5}
}
    }}
    \caption{Selective context on the introduction of \texttt{https://arxiv.org/abs/2303.07352}}
    \label{fig:appendix-case}
\end{figure*}

\begin{figure*}
    \centering
    \fbox{\parbox[c]{\textwidth}{
    {\setlength{\fboxsep}{-1pt}
    \input{tex_file/text_6}
}
    }}
    \caption{The resulting context}
    \label{fig:appendix-resulting}
\end{figure*}

\begin{figure*}
    \centering
    \fbox{\parbox[c]{\textwidth}{
    \textbf{Given Original Context}, \texttt{CUDA Memory = 77,695 MB; Time = 110.8 ms/token}

    \vspace{0.1in}
    \input{tex_file/text_3}
\ref{sec:appendix}.
    }}
    \caption{The reference summary generated by \texttt{vicuna-13b} given the full context.}
    \label{appendix-reference-summary}
\end{figure*}

\end{document}